**About the creation of a parallel bilingual corpora of web-publications**

D.V. Lande and V.V. Zhygalo
ElVisti Information center, Kiev, Ukraine

*The algorithm of the creation texts parallel corpora was presented. The algorithm is based on the use of "key words" in text documents, and on the means of their automated translation. Key words were singled out by means of using Russian and Ukrainian morphological dictionaries, as well as dictionaries of the translation of nouns for the Russian and Ukrainianlanguages. Besides, to calculate the weights of the terms in the documents, empiric-statistic rules were used. The algorithm under consideration was realized in the form of a program complex, integrated into the content-monitoring InfoStream system. As a result, a parallel bilingual corpora of web-publications containing about 30 thousand documents, was created*

### 1. Introduction

The algorithms of singling out so-called "key words" have important functions in both theory and practice. Many algorithms of singling out key words are based in vector representation and they use statistic properties of the texts. Frequency word lists in one or several languages are mostly used in singling out key words (or word bases).

The creation of a frequency word list based on the morphological dictionary (MD) using a text corpora of the documents is described in this paper, as well as the development of the algorithm of singling out key words with the use of frequency MD and a well-known approach TF IDF [1].

Based on the analysis of the key words, which were automatically singled out, and their translation into another language, the procedure of identification of duplicate documents, presented in various languages, was realized.

As it is well-known, at present the task of creating multilingual parallel text bodies is very relevant [2-4]. A suggested approach made it possible to create a bilingual Ukrainian-Russian parallel corpora of the texts from web-publications in Russian and Ukrainian languages. According to the experts, the estimated accuracy of the suggested algorithm is 98 %.

### 2. Description of the algorithm

The following procedures were used to create the parallel texts corpora:
- development of MD;
- creation of frequency dictionaries on the basis of existing MDs;
- creation of the translation dictionaries;
- realization of the algorithm of singling out key words in the document;
- translation of key words of the document into another language;
- realization of the algorithm of identifying duplicates based on the analysis of key words and their translations.

#### *2.1. Making morphological dictionaries*

Available electronic dictionaries were taken for Russian and Ukrainian languages (ispell with over 1.102 thousand word forms in the Ukrainian language and Zalizniak's dictionary which enumerates 93 392 words in a regular form).

Morphological dictionaries were supplemented by the names of well-known companies and people which were not presented in the initial dictionaries.

### 2.2. Making a frequency dictionary

To identify key words, there is a need in a frequency dictionary, which has information about each word, namely, how many times it is mentioned in some large information array and the number of documents where this word is found.

To make a frequency dictionary, a corpora of documents for the year of 2007, scanned form the Internet with help of InfoStream content-monitoring system, is taken [5]. The corpora consists of the texts from web-publications in Ukrainian (1 344 086 documents) and those in Russian (2 399 367 documents).

To instruct a frequency dictionary with help of a machine, word forms were singled out from each document, and then they were adjusted to a standard form (with certain probability). The number of both word forms and standard forms in the documents were calculated, the number of documents with this word form and/or standard from were calculated as well.

To make the retrieval of key words efficient, only those words which occurred more than twice were included in the dictionaries compiled. In addition, it was decided to use only nouns.

### 2.3. Making translation dictionaries

In the context of this research, only Russian-Ukrainian and Ukrainian-Russian dictionaries were used. Initial information for making translation dictionaries was received by means of translating nouns in standard forms using existing programs for text translation.

In the situation when one word had several translations, the most frequently used meaning was chosen which corresponded to the concept of a frequency dictionary.

### 2.4. Retrieval algorithm of key words

For the retrieval of key words, a standard approach TF IDF was used, to be more exact, its modification Okapi BM25 [6]:

$$score(D,Q) = \sum_{i=1}^{n} IDF(q_i) \cdot \frac{f(q_i,D) \cdot (k_1 + 1)}{f(q_i,D) + k_1 \cdot \left(1 - b + b \cdot \frac{|D|}{avgdl}\right)},$$

where $f(q_i,D)$ - term frequency $q_i$ in a document $D$, $|D|$ - length of the document $D$ (the number of words), $avgdl$ - an average length of a document in a corpora, $k_1$ and $b$ - free parameters, usually chosen as $k_1 = 2.0$ and $b = 0.75$. $IDF(q_i)$ − inverse frequency of a document, which is calculated with help of a formula:

$$IDF(q_i) = \log \frac{N - n(q_i) + 0.5}{n(q_i) + 0.5},$$

where $N$ - total number of documents in a corpora, $n(q_i)$ - the number of documents containing term $q_i$.

To solve the problem of homonymy (to disambiguate), it was decided to take a standard form, the most frequent one in a document corpora.

Then all the key words were ranged for each document, and the first twelve were attributed to the document. Besides, key words were automatically translated and also attributed to the document under consideration.

To improve the work of the algorithm, stop-dictionaries, which separated undesirable words, were used for each language.

According to expert estimation, 99% of quality when translating key words was reached.

## 3. Results

As a result of the research done, a new mechanism of duplicate retrieval was integrated into a content-monitoring InfoStream system, which allowed finding document duplicates in a large corpora of information with help of key words. To realize this mechanism, only 5 key words of one document, the length of which exceeded 1000 characters, had to be added to the composition of 12 key words (or their translations) of another document.

Parallel Ukrainian-Russian bodies of documents were created based on the suggested algorithm. Web-publications received for three months with help of InfoStream system were initial data for the development of the corpora (3 135 279 documents in Russian and 425 293 – in Ukrainian).

Additional criteria of separating non-completed duplicates in different languages were used:
- Total number of words in a translated version should not differ by more than 10%;
- Total number of words beginning with a capital letter (not at the beginning of a line) should not differ by more than 3 words, as the name of another information source can be put into a document;
- The quantity of numbers should not differ by more than two;
- The found number in documents should not differ by more than 15 %.

As a result of the experiment, a parallel corpora from 29 884 documents of different length was received, the accuracy of the translation being 98% (according to expert estimation).

Selected parallel bodies of the documents are available in the Internet at the address: http://www.infostream.ua/ling. The information is presented in KOI8-U, in an archived/zipped form (gzip). The total volume of the archived/zipped bodies is – 40 Mbites.

The use of this corpora for the purpose of research and teaching is free.